\documentclass[manuscript,screen,table]{acmart}

\usepackage{enumitem}
\usepackage{multirow}
\usepackage{graphicx}
\usepackage{balance}
\usepackage{caption}
\usepackage{hyperref}
\usepackage{subcaption}
\usepackage{makecell}
\usepackage{etoolbox}
\usepackage{xcolor}

\newcommand{\hlbox}[2]{
  \begin{center}
    \fcolorbox{white}{white}{
      \parbox{0.9\columnwidth}{\noindent \textbf{#1}. \textit{#2}}
    }
  \end{center}
}
\AtBeginDocument{%
  }

\copyrightyear{2023}
\acmYear{2023}
\setcopyright{acmcopyright}
\acmConference[LAK 2023]{LAK23: 13th International Learning Analytics and Knowledge Conference}{March 13--17, 2023}{Arlington, TX, USA}
\acmBooktitle{LAK23: 13th International Learning Analytics and Knowledge Conference (LAK 2023), March 13--17, 2023, Arlington, TX, USA}
\acmPrice{15.00}
\acmDOI{10.1145/3576050.3576148}
\acmISBN{978-1-4503-9865-7/23/03}

\begin{document}

\title[Unknown Unknowns to Student Success Predictors in Online-Based Learning]{Do Not Trust a Model Because It is Confident: Uncovering and Characterizing Unknown Unknowns to Student Success Predictors in Online-Based Learning}

\author{Roberta Galici}
\affiliation{
 \institution{University of Cagliari}
 \streetaddress{Department of Mathematics and Computer Science}
 \city{Cagliari}
 \state{Sardinia}
 \country{Italy}
 \postcode{09124}}
 \email{roberta.galici@unica.it}

\author{Tanja Käser}
\affiliation{
  \institution{EPFL}
  \streetaddress{EPFL IC IINFCOM ML4ED INF Building – Station 14}
  \city{Lausanne}
  \state{Vaud}
  \country{Switzerland}
  \postcode{CH-1015}}
  \email{tanja.kaser@epfl.ch}

\author{Gianni Fenu}
\affiliation{
 \institution{University of Cagliari}
 \streetaddress{Department of Mathematics and Computer Science}
 \city{Cagliari}
 \state{Sardinia}
 \country{Italy}
 \postcode{09124}}
 \email{fenu@unica.it}

\author{Mirko Marras}
\affiliation{
 \institution{University of Cagliari}
 \streetaddress{Department of Mathematics and Computer Science}
 \city{Cagliari}
 \state{Sardinia}
 \country{Italy}
 \postcode{09124}}

\begin{abstract}
Student success models might be prone to develop weak spots, i.e., examples hard to accurately classify due to insufficient representation during model creation.
This weakness is one of the main factors undermining users' trust, since model predictions could for instance lead an instructor to not intervene on a student in need.
In this paper, we unveil the need of detecting and characterizing unknown unknowns in student success prediction in order to better understand when models may fail. 
Unknown unknowns include the students for which the model is highly confident in its predictions, but is actually wrong. 
Therefore, we cannot solely rely on the model's confidence when evaluating the predictions quality.
We first introduce a framework for the identification and characterization of unknown unknowns. 
We then assess its informativeness on log data collected from flipped courses and online courses using quantitative analyses and interviews with instructors.
Our results show that unknown unknowns are a critical issue in this domain and that our framework can be applied to support their detection.
The source code is available at \href{https://github.com/epfl-ml4ed/unknown-unknowns}{https://github.com/epfl-ml4ed/unknown-unknowns}. 
\end{abstract}




\maketitle

\section{Introduction}

\vspace{2mm} \noindent \textbf{Context and Objective}. 
Learning analytics has been fuelled through increased access to \emph{data} and analytical studies development \cite{DBLP:journals/widm/RomeroV20}.  
These studies have often focused on factors that influence \emph{student success} to bring insights into the learning process and establish recommendations.
Student success has been proven to be influenced by many factors, such as 
engagement \cite{cho2013self,dogan2015student}, 
regularity \cite{DBLP:journals/ce/JovanovicMGDP19,DBLP:conf/ectel/BoroujeniSKLD16}, 
critical thinking \cite{vzivkovil2016model}, 
metacognition \cite{hrbavckova2012relationship,narang2013metacognition}, 
and socio-emotional well-being \cite{berger2011socio,toscano2020emotional}.
This knowledge has been capitalized to build \emph{student models} that can predict course success \cite{DBLP:journals/umuai/GardnerB18,alyahyan2020predicting}.
Their predictions are useful for several means, e.g., personalized intervention, adaptive content provisioning, and learning process understanding. 
Notable instructional strategies benefiting from this support include \emph{blended learning} (e.g., flipped courses \cite{DBLP:journals/caee/Rubio-Fernandez19}) 
and \emph{online learning} (e.g., MOOCs \cite{DBLP:conf/t4e/MSCB16}),
where instructors can only partially observe student learning. 
However, these models also introduce risks for students and instructors. Their users may not be used to reasoning about the \emph{forms of model uncertainty} and thus \emph{misunderstand or mistrust} its predictions. 
Trust is essential for the acceptance of these models in education 
\cite{DBLP:journals/bjet/NazaretskyACA22,DBLP:conf/lak/TsaiG17,nazaretsky2022instrument}.
Hence, our goal is to investigate \emph{weak spots} of student success models to understand the circumstances under which 
these \emph{models should (not) be trusted} by the users.

\vspace{2mm} \noindent \textbf{Open Problem}. 
Humans are often convinced that predictions should be trusted, as \emph{they result from an accurate machine-learning model}
\cite{DBLP:journals/corr/abs-1907-12652,DBLP:conf/fat/ZhangLB20,DBLP:conf/chi/YinVW19}. 
By accurate we mean that if the model predicted well in the past it should be trusted to do so in the future. 
However, there is less evidence on the \emph{weak spots} these models can develop, i.e., examples hard to accurately classify due to low representation. 
Our study focuses on a weak spot never explored in student success models: \emph{unknown unknowns} \cite{DBLP:journals/jdiq/AttenbergIP15}. 
They denote students for which the model is highly confident of its prediction although it is wrong\footnote{
On the contrary, \emph{known unknowns} are students for which the model is unsure about the correct prediction, since it falls close to the decision boundary.
This phenomenon has been studied in the active learning field, and a solution could be to solicit new labels from humans on uncertain examples.}. 
Syntactically, the second "unknowns" represents model knowledge (does not know correct labels), and the first "unknown" represents model awareness (not aware of being wrong; very confident of the wrong predictions).
Unknown unknowns might result from unmodeled data biases (e.g., under-represented groups), 
data distribution shifts (e.g., course changes),
or factors hidden to the model (e.g., students' prior knowledge).
Efforts to analyze unknown unknowns in fields other than education have uncovered \emph{unintended machine behaviours}
\cite{DBLP:conf/kdd/ShabatSA17,DBLP:conf/aaai/0006ZZ21,liu2020towards}.

\begin{figure*}[!t]
\centering
\includegraphics[width=1.0\linewidth, trim=4 4 4 4,clip]{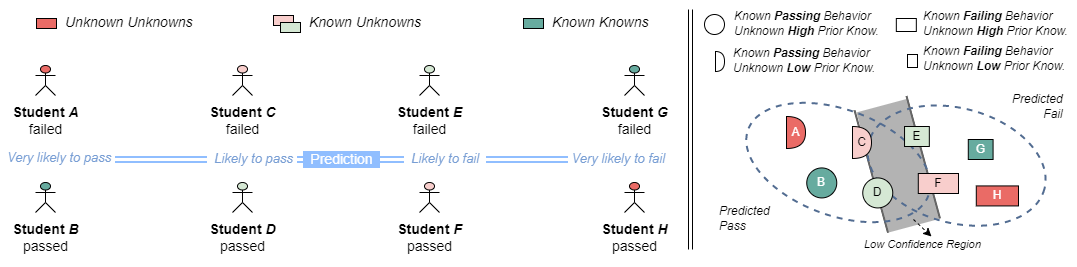}
\caption{\textbf{Motivating Example}. 
On the left, we see a student success model highly confident, but actually wrong, on predictions for students \texttt{A} and \texttt{H} (unknown unknowns),
both highly confident and correct on students \texttt{B} and \texttt{G} (known knowns), and only slightly confident of its predictions on students \texttt{C}, \texttt{D}, \texttt{E}, and \texttt{F} (known unknowns).
On the right, we assume that students' prior knowledge was a variable the model was not aware of.
We illustrate that not only students' behavior (visible to the model), but also students' prior knowledge (outside of the scope of the model) was relevant for success. Student \texttt{A} exhibited a typical passing behavior, but had a low prior knowledge. On the contrary, Student \texttt{H} exhibited a typical failing behavior, but had a high prior knowledge.}
\label{fig:toy-example}
\vspace{3mm}
\end{figure*}

\vspace{2mm} \noindent \textbf{Motivating Example}. 
The toy example in Figure \ref{fig:toy-example} shows the \emph{unknown unknowns} problem. \emph{Student A} is a 1st year university student. The student failed a Math course taught in a flipped format. During the course, they had access to a platform for completing pre-class activities, e.g., watching videos and doing exercises. We assume that a machine-learning model flagging students needing intervention (based on educationally relevant features) was used in this course. This model was built upon historical pre-class behavioral data from the same course and demonstrated a high accuracy. However, this data included a majority of students with high prior knowledge in Math. Due to this, the model found that a low time spent on videos (a feature) was predictive of success (e.g., content was often confirmatory knowledge). \emph{Student A} did not spend much time on videos because they were struggling. The model, unaware of students' prior knowledge, indicated to the instructor \emph{Student A} was very likely to pass the course. Due to the limited teaching resources, the team decided to not intervene on \emph{Student A}, who ended up failing the course. In this case, students' prior knowledge led to unknown unknowns. 

\begin{figure*}[!b]
\centering
\vspace{3mm}
\includegraphics[width=.9\linewidth, trim=4 4 4 4,clip]{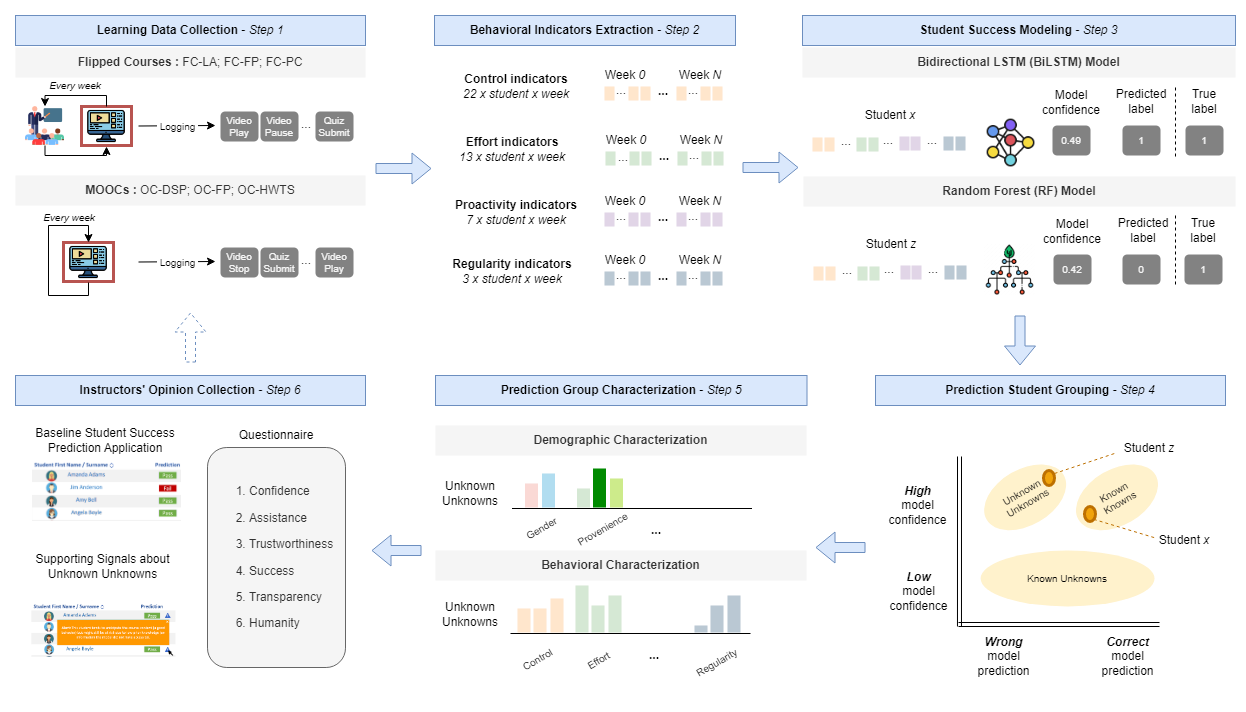}
\vspace{-5mm}
\caption{\textbf{Methodology}. 
We collected log data from flipped courses and MOOCs (Step 1) and extracted behavioral indicators of control, effort, proactivity, and regularity (Step 2). 
We built a random forest classifier and a bidirectional LSTM classifier, both returning a confidence score and a predicted label (Step 3).
We then grouped students into known knowns, known unknowns, and unknown unknowns based on the confidence score, predicted label, and original true label (Step 4). 
The three groups were characterized to identify (dis)similarities (Step 5). Finally, we asked instructors for their opinion and support on unknown unknowns detection (Step 6).
} 
\label{fig:methodology}
\end{figure*}

\vspace{2mm} \noindent \textbf{Our Contribution}. 
In this paper, we unveil the need of characterizing unknown unknowns in student success prediction to gain a deeper understanding of when the model fails. 
Bridging education, trust, and machine learning concepts, we introduce a computational framework for cost-efficient identification and characterization of unknown unknowns. 
We then assess its effectiveness, informativeness, and cost-efficiency on two state-of-the-art student success prediction models, 
based on data from six courses (three flipped classroom courses and three MOOCs). 
With our experiment, we investigate 
(i) \emph{whether unknown unknowns exist in student success prediction and how they vary in number and type across flipped courses and MOOCs}, 
(i) \emph{whether we can characterize unknown unknowns in student success prediction models}, and 
(iii) \emph{the extent to which providing instructors with information about unknown unknowns impact their perception of the student success model}. 
Our results show that unknown unknowns are a serious issue for student success models and
highlight the value of our framework for the discovery of unknown unknowns across experimental conditions.

\section{Methodology}
In this section, we describe the educational scenario (learning activities, tracked data, behavioral indicators of interest) as well as 
the student success models (model creation, evaluation, students grouping, group characterization, instructors' opinions) leveraged for unknown unknown analysis. 
Figure \ref{fig:methodology} provides an overview of the steps of our framework. 

\subsection{Learning Data Collection - Step 1 / 6}
While the characterization of unknown unknowns may be important for any educational scenario supported by student success models, our study focused on two teaching strategies applied to courses within a European university: flipped classroom courses and MOOCs. 
In the following, we explain the teaching scenario for both course types in more detail.

\begin{table*}[!t]
\vspace{3mm}
\caption{\textbf{Learning Data Collection}. Detailed information about the courses included in our study.}
\label{tab:courses}
\vspace{-3.5mm}
\small
\resizebox{\textwidth}{!}{
\begin{tabular}{lllrrllrrr}
\toprule
\textbf{Course Title} & \textbf{ID} & \textbf{Field$^1$} &  \textbf{Setting} & \multicolumn{1}{r}{\textbf{\begin{tabular}[c]{@{}c@{}} Students$^2$\end{tabular}}} & \textbf{Level} & \textbf{Language} & \multicolumn{1}{r}{\textbf{\begin{tabular}[c]{@{}c@{}} Weeks\end{tabular}}} &  \multicolumn{1}{r}{\textbf{\begin{tabular}[c]{@{}c@{}} Failing Rate \end{tabular}}} & \multicolumn{1}{r}{\textbf{\begin{tabular}[c]{@{}c@{}} Quizzes\end{tabular}}} \\
\midrule
Linear Algebra & \texttt{FC-LA} & Math & Flipped & 292 & BSc & English & 14 & 40.00\% & 179 \\
Functional Programming & \texttt{FC-FP} & CS & Flipped & 216 & BSc & French & 18 & 38.53\% & 0 \\
Parallelism and Concurrency & \texttt{FP-PC} & CS & Flipped & 147 & MSc & French & 16 & 37.16\% & 0 \\
\hline
\hline
Digital Signal Processing & \texttt{MOOC-DSP} & CS & MOOC & 15,394 & MSc & English & 10 & 75.71\% & 38 \\
Household Water Treatment and Storage & \texttt{MOOC-HWTS} & NS & MOOC & 2,423 & BSc & French & 6 & 47.36\% & 10 \\
Functional Programming Principles in Scala & \texttt{MOOC-FP} & CS & MOOC & 18,702 & BSc & French & 8 & 42.15\% & 3 \\
\bottomrule
\end{tabular}}
\scriptsize{$^1$ \textbf{Field}: \textit{CS}: Computer Science; \textit{Math}: Mathematics; \textit{NS}: Natural Science}. \; \; 
\scriptsize{$^2$ \textbf{Students}: for MOOCs, number of students obtained after removing early-dropout students \cite{DBLP:conf/lats/SwamyMK22}.}
\vspace{2mm}
\end{table*}

\vspace{2mm} \noindent \textbf{Learning through Flipped Classroom Courses}.
We based our analysis on three semester-long university courses with a weekly schedule including both lectures and sessions of recitation or exercises. 
These courses followed the learning science principles detailed in \cite{hardebolle2022gender}.
All courses illustrated in Table \ref{tab:courses} are compulsory for Computer Science and Communication Systems Bachelor's degrees at said European university. 
In class, activities included quizzes, short problem-solving exercises, and structured proof-type problems. 
In addition, students were expected to spend few hours per week on individual study as a preparation for class (referred to as pre-class activities). 
Instructions regarding the preparatory work were sent to students (a list of sections from a MOOC with video lectures and online quizzes) one week in advance. 
Quizzes, usually including multiple-choice questions, enabled students to self-assess their learning. 

The MOOC platform collected data about student pre-class activities for all three flipped courses. 
Log entries were structured to report the user, activity, and timestamp (e.g., user: 10, activity: play video 32, timestamp: 05-03-2018 12:06:01). 
These entries were complemented with demographic attributes about students, including their gender, geographic provenience, and high school diploma. 
No data about in-class activities was recorded.
Student achievement was measured by their grade in the final exam (grade between 1 and 6, with 4 or higher indicating a passing grade). 
This study was approved by the university’s ethics committee (HREC 058-2020/10.09.2020, 096-2020/09.04.2020).

\vspace{2mm} \noindent \textbf{Learning through Massive Open Online Courses}.
We complemented our analysis with three MOOCs taught by three different instructors of the same European university in Coursera (Table \ref{tab:courses}, last three rows). 
These courses were accessible to all people worldwide. 
Lectures were released on a weekly basis and students were expected to spend some hours per week to complete them. 
Each week, courses included short video lectures (10-15 minutes) introducing key concepts and quizzes for self-assessment. 
Students were asked weekly to complete a graded assignment, their score was used by the instructor to judge student achievement.
The final course grade ranged between 0 and 100 (with at least 60 points to pass) and was calculated based on a weighted sum of the weekly assignment scores plus a final exam.

We collected over 145,640 log entries, including the user, the activity, and the timestamp, reported in the same format as for flipped course logs. 
Students' gender and geographic provenience were attached to the log data where students provided them voluntarily. 
Collected data included all the activities performed by the student in the platform, covering a large part of their learning (except, for instance, offline video watching).
In the MOOC setting, our study focused on a scenario where student success models were built based on the entire activity data in the platform (thus not visible to the instructor). Student achievement was measured by using the final course grade.

\subsection{Behavioral Indicators Extraction - Step 2 / 6}\label{sec:features-descr}
Prior work has found significant association with academic achievement for self-regulated learning (SRL) aspects \cite{cho2013self,DBLP:conf/lak/SherHG20,DBLP:conf/aied/Mejia-Domenzain22}: 
effort regulation (persistence in learning), 
time management (ability to plan study time), 
metacognition (awareness and control of thoughts), 
critical thinking (ability to carefully examine material), 
and help-seeking (obtaining assistance if needed). 
A variety of learning indicator sets have been proposed in the literature accordingly. 
Their power for modeling learning success has been compared by \cite{DBLP:conf/edm/MarrasVK21}, which identified the most important ones in both flipped courses and MOOCs.
Empirical evidence of their importance across MOOCs was also provided by \cite{DBLP:conf/lats/SwamyMK22}. 
Both our logging policy and experimental scenario were similar to those adopted in prior work. 
Thus, we built our models on top of the indicators proven to be important to them (see Table 3 in \cite{DBLP:conf/lats/SwamyMK22} for a full list and description). 
The granularity and comprehensiveness of the collected log data allowed the studies mentioned above (and our) to consider 
effort regulation (effort), time management (regularity and proactivity), and metacognition (control) dimensions.
Critical thinking and help-seeking could not be measured. 
The considered learning indicators covered the following dimensions. 

\vspace{2mm} \noindent \textbf{Control} 
(22 indicators per student per week) models in-video and cross-video behavior as a proxy of student ability to control the cognitive load of video lectures
through course weeks (metacognition). 
For instance, in-video flow, manageable through the platform functionalities (e.g., pause button), could include regular pauses to segment learning
\cite{hrbavckova2012relationship,narang2013metacognition}. 
Among others, this feature set consists of the proportion of videos watched, re-watched, or interrupted.

\vspace{2mm} \noindent  \textbf{Effort} 
(13 indicators per student per week) aims at monitoring how much and how frequently students engage with the course content (both videos and quizzes), 
proven to be fundamental for learning success \cite{cho2013self,dogan2015student}. 
These features included indicators such as the total number of student clicks on weekends and on weekdays, and the total number of sessions.

\vspace{2mm} \noindent  \textbf{Proactivity} 
(7 indicators per student per week) attempts to measure the extent to which students are on time or ahead of the schedule,
proven to be prediction of performance especially in MOOCs \cite{DBLP:conf/edm/MarrasVK21}. 
These features are related to completion of videos and quizzes, according to the week of the course they are schedule on. 
Example features included the number of scheduled videos watched for that week and the number of quizzes passed on the first try.

\vspace{2mm} \noindent \textbf{Regularity} (3 indicators per student per week) is also associated with time management. 
It estimates the intra-week and intra-day time management patterns (i.e., capturing whether a student regularly engages on specific weekdays or day times),
which have been proved to be predictive of student success in MOOCs and flipped classrooms \cite{DBLP:journals/ce/JovanovicMGDP19,DBLP:conf/ectel/BoroujeniSKLD16}. 

\vspace{2mm} Since indicator scores vary in their range, we performed a min-max normalization per feature across all students and weeks for that feature. 
Though the selected indicators fall into four dimensions and include several measures per dimension, 
we acknowledge that other relevant dimensions and measures could be beneficial for student success modeling. 
Our study can be easily run on a different (and larger) set of behavioral indicators.

\subsection{Student Success Modelling - Step 3 / 6}
A wide range of student success models have been proposed so far in the literature \cite{DBLP:journals/umuai/GardnerB18}.
To align with prior work, since our study does not aim to propose a novel model, 
we considered two models reporting a high accuracy while providing a certain level of interpretability. 
Random Forest (RF) classifiers have achieved this in both flipped and MOOC contexts, when fed with behavioral features 
\cite{DBLP:conf/lak/GoidsenhovenBDB20,DBLP:conf/edm/MarrasVK21}.
Recent neural network classifiers based on BiLSTMs including attention layers, sigmoid activation, and a cross entropy loss function, have resulted in higher accuracy \cite{DBLP:conf/lats/SwamyMK22} and good interpretability as well \cite{swamy2022evaluating}.
Again, we based our decision on the similarity of the underlying context and logging system. 
We acknowledge that other models, e.g., Linear Regression and Support Vector Machines, have been used in prior works (e.g., \cite{DBLP:journals/ce/JovanovicMGDP19}), but we left their analysis as a future work, using RFs as a representative of this class of models.

\begin{figure*}[!t]
\centering
\includegraphics[width=.8\textwidth, trim=4 4 4 4,clip]{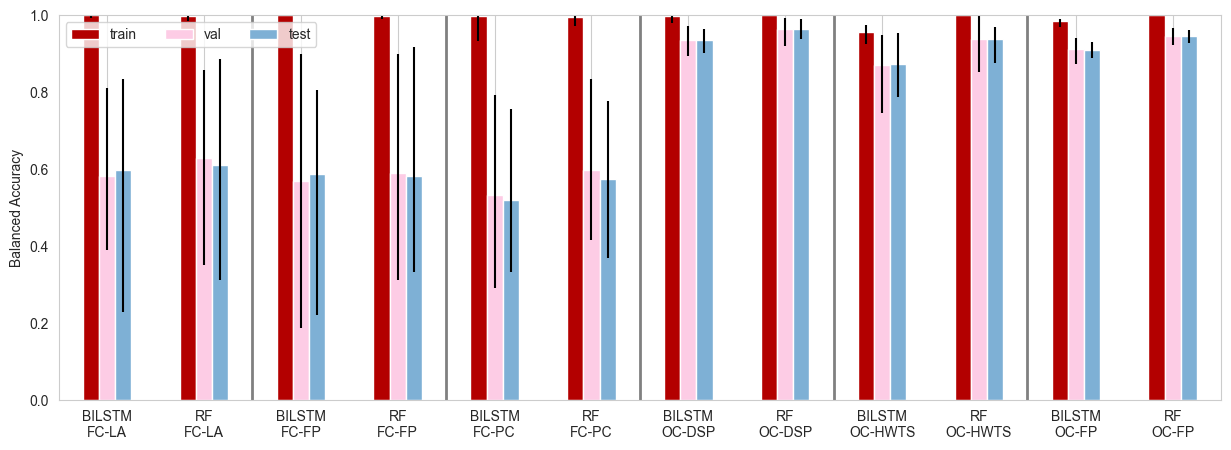}
\vspace{-3mm}
\caption{\textbf{Student Success Models Performance}. Error bars reporting the average, maximum, and minimum balanced accuracy obtained by random forest
and bidirectional LSTM models on a nested 10-fold cross-validation for each of the courses in our study.} 
\label{fig:ssm-bacs}
\vspace{2mm}
\end{figure*}

For each course and model, we applied a nested student-stratified (i.e., dividing the folds by students) 10-fold cross validation. 
The same folds were used for all experiments across models, and we optimized the hyper-parameters via grid search.
In each iteration, we ran an inner student-stratified 10-fold cross-validation on the training set in that iteration, 
and selected the combination of hyper-parameter values yielding the highest accuracy on the inner cross-validation. 
Finally, 200 models per course were obtained ($2$ architectures $\times \; 10$ outer folds $\times \; 10$ inner folds).
For each model, the balanced accuracy was computed on the training and validation to check validity and on the test set to assess model generalizability. 
In other words, we pulled a random set of students out of the course and demonstrate generalizability in the same context.
For conciseness, we do not report other similar evaluation metrics, such as kappa score and area under the ROC curve. 
Figure \ref{fig:ssm-bacs} collects the averaged balanced accuracy on the three sets for each course and architecture combination. 
Models predicting on MOOCs provided a higher balanced accuracy than those on flipped courses. 
This might be justified by the fact that online pre-class activities in flipped courses are a tiny portion of the learning process. 
RF and BiLSTM often performed equally well, with the former reporting a slightly higher balanced accuracy on MOOCs.

Each model was able to predict the probability $p$ that the student would fail the course, as an output. 
Specifically, $p$ could range between $0$ and $1$, with values closer to $1$ meaning that the student would probably fail the course. 
We finally used a decision threshold equal to $0.5$ to obtain the final predicted label:
$\Tilde{y}=0$ in case $p < 0.50$ (predicted pass); 
$\Tilde{y}=1$ in case $p \geq 0.50$ (predicted failure). 
This threshold is a common value in general machine learning and prior education-related works.
The model confidence was obtained by measuring the extent to which the probability $p$ was close to the decision threshold.
Numerically, the model confidence $c$ for a predicted probability $p$ is equal to $| p - 0.50 |$, ranging between $0$ and $0.50$. 
Higher values mean that the model is more confident about the prediction. 

\subsection{Prediction Student Grouping - Step 4 / 6} \label{sec:grouping}
Obtained a trained student success model, computing an importance score for each indicator and link it to successful patterns has been a common practice concerning both RFs
\cite{DBLP:conf/edm/MarrasVK21} 
and linear regression models
\cite{DBLP:journals/ce/JovanovicMGDP19}. 
Recent work has also extracted these scores through explainability methods \cite{swamy2022evaluating}. 
These works however assume that the model accurately reflects the relationships between learning and students' success,
which might not be true for a (non-negligible) portion of students. 
Since no student should be negatively impacted by a model, such portion of students must not be ignored.  

\begin{figure*}[!t]
\centering
\includegraphics[width=.8\textwidth, trim=4 4 4 4,clip]{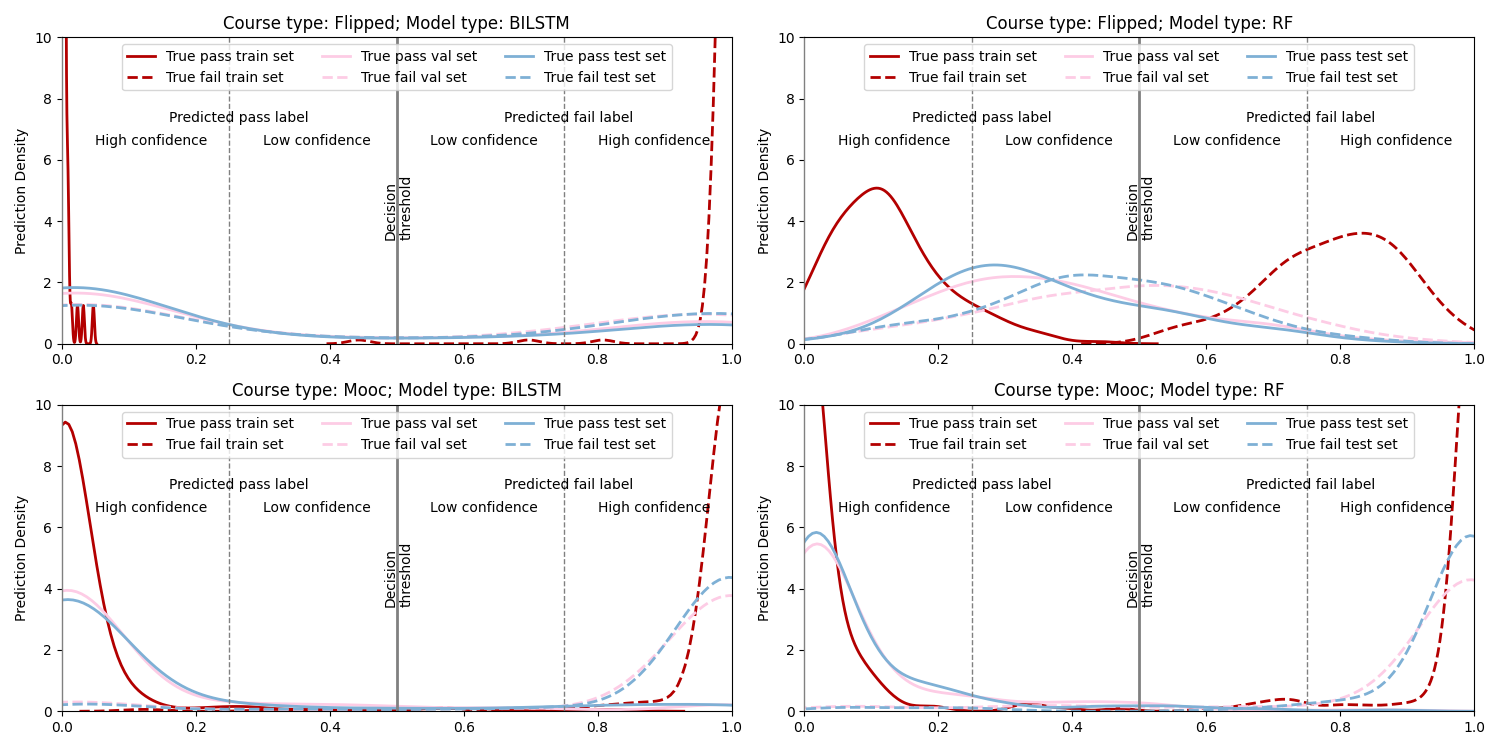}
\caption{\textbf{Student Success Predicted Probabilities}. Distributions of predicted failing probabilities for students belonging to the training, validation,
and test sets, under the four combinations of course type (Flipped or MOOC) and model type (RF or BiLSTM).} 
\label{fig:ssm-scores}
\vspace{2mm}
\end{figure*}

To illustrate an emerging issue, see Figure \ref{fig:ssm-scores}. 
Each plot collects the predicted probability distribution for students within training, validation, and test sets,
considering $200$ pass and $200$ failing students selected randomly. 
On flipped courses, BiLSTM models were often very confident about their predictions, being skewed at the two horizontal extremes. 
However, looking at Figure \ref{fig:ssm-bacs}, their balanced accuracy on flipped courses was not high.
Concerning MOOCs (bottom row), the same models were again very confident, but in this case their accuracy was very high. 
Interestingly, on flipped courses, RF showed a high uncertainty in predictions for unseen validation and test students. 
These results raise concerns on the possibility the model might be very confident but actually wrong and, in turn, open to severe consequences (e.g., deviating humans' understanding of the learning process) in case models are used in the real world.

Given the above observations, investigating the interplay between the predicted label correctness and the model confidence becomes essential.   
Let assume that $\delta \in (0, 0.50)$ is the hypothetical trust level the user has while using model predictions. 
Predictions with $c \geq \delta$ would be considered by the user with a higher trust. 
Conversely, the final users would consider those having $c < \delta$ with a lower trust. 
Given a student known to have a true label $y$ and a model that predicts a label $\Tilde{y}$ with a confidence $c$, 
the student can be assigned to a group $g \in \{0, 1, 2\}$ as follows:

\begin{itemize}
\item \textbf{Known Knowns} ($g = 0$: $c \geq \delta$ and $\Tilde{y} = y$) include students for whom 
the model is very confident and the corresponding predictions correctly reflect the true success or failure.  
These are the ideal cases where the model is correctly optimistic: 
the model reports high confidence and correct label for the examples in these region(s). 
\item \textbf{Known Unknowns} ($g = 1$: $c < \delta$ and any $\Tilde{y}$) include students for whom 
we are aware the model is not sure and the corresponding predictions would require extra care before being used to take action in the real world. 
This concept models cases for which errors are expected based on the probability estimates of the classification.
\item \textbf{Unknown Unknowns} ($g = 2$: $c \geq \delta$ and $\Tilde{y} \neq y$) include students for whom 
the model is confident but their predictions are actually wrong. 
Intuitively, these are examples distant from the decision boundary yet labeled incorrectly. 
These examples may be rare to be detected, but their non-negligible prevalence makes them a risk. 
\end{itemize}

For convenience, we used a trust level $\delta = 0.25$ and left analyses under other trust levels as a future work\footnote{
This threshold let us provide a more focused unknown unknowns comparison, but is not required by the final system. 
In any case, it might be adjusted according to the model (e.g., its accuracy).
For convenience, our study considers the same threshold for both models to make the comparison fair.  
}. 

\subsection{Prediction Group Characterization - Step 5 / 6} \label{sec:explanatory}
To reduce unknown risks, someone could think that being able to classify a future student as unknown unknown would be the solution. 
This might mean to train a second model to predict whether a student is an unknown unknown and do something with it. 
This solution unfortunately has several weaknesses, e.g., complexity, more biases, and lack of data. 
If another model can detect unknowns unknowns, that model could be directly used to improve the original student success model. 
Such weak spot however comes due to knowledge that model (and the second model) was missing. 

Letting stakeholders be aware of unknown unknown cases would conversely provide a better support.
Typical archetypes of unknown unknowns could be provided to the instructors before or while using the model.  
Indeed, a human might think about the information the model was missing and propose to fix data collection and processing accordingly. 
In this scenario, a core question is the choice of the variables used to explain the unknown unknown cases. 
Two complementary approaches exist for choosing them \cite{nilsen2020exploratory}. 
In confirmatory research, the potential impact of different variables is hypothesized a-priori, based on existing theories. 
An exploration-driven approach is used when there is a lack of theories or while generalizing across domains. 
Given the only recently awareness of unknown unknowns, exploration research could produce new hypotheses to be evaluated later by experts (e.g., instructors) in their courses.

To characterize unknown unknowns from the model perspective, we studied the relationship between
(i) model confidence and predicted label correctness according to the prediction-based student groups we identified in the previous step, and
(ii) a range of variables the model already knows (i.e., the behavioral indicators leveraged for training).
To characterize unknown unknowns, as a source of context, we also considered some variables pertaining to demographics.
Not included in the model training (also for ethical reasons), demographics might be available and used as a set of contextual variables.  
Formally, the \emph{independent variables} for our characterization were represented, for each student, with a vector $v$ including 
the $45$ behavioral indicators (averaged across course weeks) and 
the $2$ demographic attributes (gender and provenience). 
The \emph{dependent variable} was the group label $g$ computed in the previous section. 

Intuitively, we were interested in identifying variables that have a statistically significant relationship with the dependent variable. 
A multiple regression model was used to analyze these relationships, fitting it with 
the vectors $v$ as an input and the corresponding group labels $g$ as an output. 
Formally $y = \epsilon + \gamma_0 + \sum_{j} \gamma_{j} \cdot v_{j}$, where  
$y$ is the dependent variable, and $\gamma_j$ and $v_{j}$ respectively represent the regression coefficient and the value of the $j$-th variable.
The coefficients $\gamma \in \Gamma$ indicate us the extent to which a change of a given variable, holding all the others constant, leads to a change in the group.
Being a multiple regression, our analysis tested the null hypothesis that all coefficients $\gamma \in \Gamma$ are zero, 
versus the alternative that at least one coefficient $\gamma_j$ is nonzero.
We conjecture that variables with a nonzero statistically significant coefficient might be deemed as important to explain the unknown unknown membership.

\subsection{Instructors' Opinion Collection - Step 6 / 6} \label{sec:quest}
Finally, we were interested in understanding how the instructors' perception of student success model would change when they are made aware of unknown unknowns. 
To this end, we designed a questionnaire\footnote{
The full questionnaire is available at the following webpage: \href{https://shorturl.at/qvX47}{https://shorturl.at/qvX47}.
}, following prior work on trust in artificial intelligence for education \cite{nazaretsky2022instrument}, 
including two main sections (see Table \ref{tab:questionnaire}): 
one on demographic information, whereas
the other was focused on a use case on unknown unknowns in student success modelling. 

In the first part, we were interested in knowing who the experts were, including the type of their organization (e.g., academia, industry), their role in the organization (e.g., full professor, researcher), the continent and country they are based in, the age group (in a specific range), and the gender identity. In addition, we asked participants how many courses they acted as an assistant (e.g., tutor) and as an instructor (e.g., full professor, associate professor) in. 

\begin{table}[!t]
\caption{\textbf{Instructors' Questionnaire}. Content of the questionnaire provided to instructors in our study.}
\label{tab:questionnaire}
\parbox{.4\textwidth}{
\centering
\footnotesize
\vspace{-5.75mm}
\begin{tabular}{ll}
    \toprule
    \textbf{ID}  & \textbf{Question on Demographics}    \\
    \midrule                                                                               
    \textit{Q01} & Which type of organization are you based in?                                                                  \\
    \textit{Q02} & Which role do you have in your organization?                                                                  \\
    \textit{Q03 (Q04)} & Which continent (country) are you based in?                                                                             \\
    \textit{Q05} & Which age group are you in?                                                              \\
    \textit{Q06} & With which gender identity do you most identify?                                                              \\
    \textit{Q07} & How many courses were you teaching assistant for?                                    \\
    \textit{Q08} & How many courses were you an instructor for? \\
    \bottomrule
\end{tabular}
}
\hfill
\parbox{.55\textwidth}{
\vspace{-3mm}
\centering
\footnotesize
\renewcommand{\thesubfigure}{ii-z}
\begin{tabular}{ll}
    \toprule
    \textbf{ID}          & \textbf{Question on Student Success Models}                                                                            \\ 
    \midrule                           
    \textit{Q09.1 (Q10.1)} & Be (less) confident about using model predictions.                                           \\
    \textit{Q09.2 (Q10.2)} & Feel assisted (more) in diagnosing students’ difficulties.                                   \\
    \textit{Q09.3 (Q10.3)} & Trust (less) using model predictions in your classroom.                                      \\
    \textit{Q09.4 (Q10.4)} & Feel (more) successful in using model predictions.                                           \\
    \textit{Q09.5 (Q10.5)} & See as (less) transparent how model predictions are made.                                    \\
    \textit{Q09.6 (Q10.6)} & \makecell[l]{Rely on model predictions at least as much as you rely on \\ a recommendation from a colleague.} \\
    \bottomrule
    \\
\end{tabular}
}
\end{table}

In the second part, we described a use case with student success predictions presented to the instructor while investigating which students would require assistance. 
This use case was accompanied by a provocative user interface which would require them to think about the influence of student success models on the instructor\footnote{ 
The user interface was on purpose limited to a rudimentary pass/fail setting, without any confidence level, 
to stimulate instructors' critical thinking and reduce the impact of other user interface elements on the unknown unknowns perception.}. 
First, we asked experts to rate their perceived confidence, assistance, success, trust, transparency, humanity with respect to this case.
This decision let us understand the original perception of instructors towards student success models in general to better contextualize our results.  
We then explained the concept of unknown unknowns and envisioned a simple addition to the user interface, indicating students at risk of being unknown unknowns via alerts and explanations.
This information would trigger instructors to question the provided prediction and investigate the corresponding students' behavior more thoroughly. 
We then asked the extent to which their perception of the student success model changed relatively to their original perception (e.g., more or less trust). 
In both series of questions, each participant was allowed to select among four possible answers, from "strongly disagree" to "strongly agree". 
No neutral question was introduced; hence,  answers should be interpreted as what the instructors would answer in case they were forced to make a decision. 
As a final field, we asked if they identified any (dis)advantages raised by being aware of unknown unknowns.

Following \cite{fenu2022experts}'s protocol, we e-mailed the questionnaire to experts with a paper accepted in a top conference in education in 2021 (AIED, EAAI, EC-TEL, EDM, ICALT, ITS, LAK, L@S). 
Out of 1,721 people, 112 (6,51\%) completed the questionnaire.
This choice was made to include people who are both educators and experts in the field.
Though we acknowledge that future work will need to focus also on the perception of a generic instructor,
unknown unknowns analysis is still at early stages and our feedback can make more mature our understanding before involving them. 

\section{Experimental Results}
Although our methodology can be used to analyze several perspectives, we focused on the following research questions: 

\begin{itemize}
\item Do unknown unknowns exist in success prediction? How do they vary across flipped courses and MOOCs?
\item Can we characterize common unknown unknowns in the considered student success prediction models?
\item How does providing instructors with signals about unknown unknowns impact their perception of the model? 
\end{itemize}

\subsection{RQ1: (Un)known Unknown Prevalence}
\begin{figure*}[!t]%
    \centering
    \hspace{0.8cm}
    \subfloat[Flipped courses\label{fig:groups-flipped}]{
    \includegraphics[width=.8\textwidth, trim=4 4 4 4,clip]{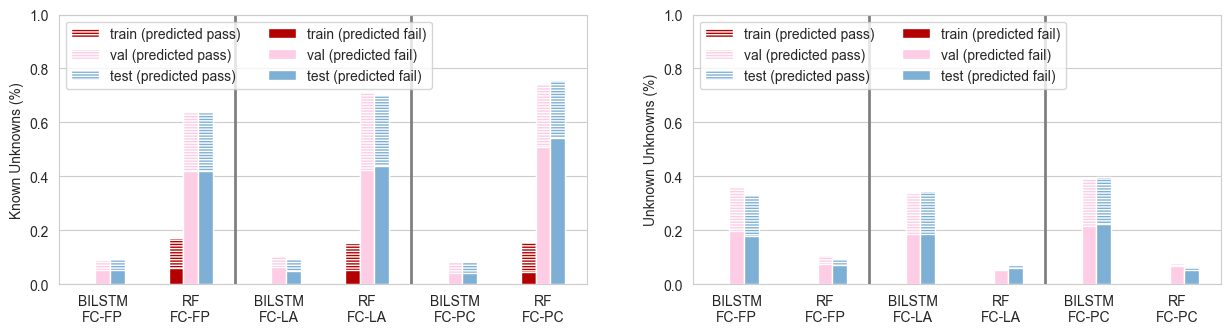}}
    \newline
    \subfloat[MOOCs\label{fig:groups-mooc}]{
    \includegraphics[width=.8\textwidth, trim=4 4 4 4,clip]{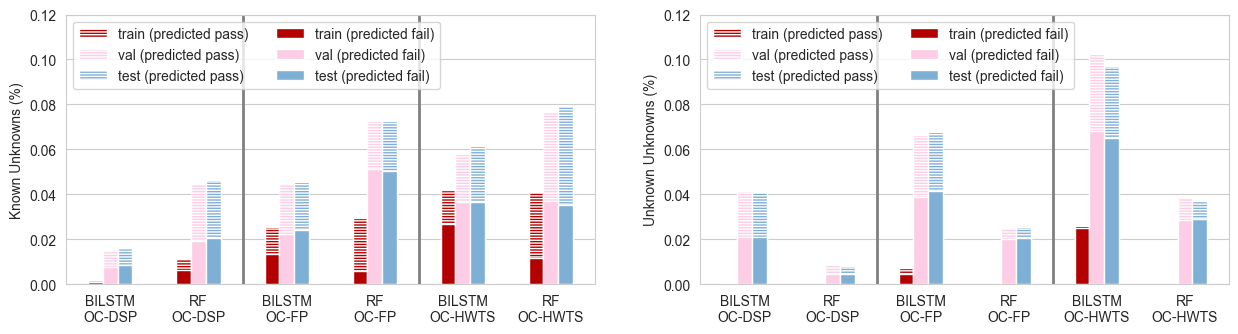}}
    \caption{\textbf{Prediction Student Groups}. 
    Average percentage of students in the training, validation, and test sets being a known unknown (left) and unknown unknown (right).
    Solid (dashed) bars indicate students who passed (failed) but were predicted as failing (passing).}
    \label{fig:groups}
    \vspace{3mm}
\end{figure*}

In a first analysis, we were interested in understanding whether (un)known unknowns exist in the developed student success models and 
in investigating whether any difference between courses of the same type and across courses of a different type arises. 
To this end, Figure \ref{fig:groups} depicts the average percentage of students belonging to these two groups across the considered settings. 
As expected from the model performance in Figure \ref{fig:ssm-bacs}, more (un)known unknowns were present in flipped courses than MOOCs. 
We considered courses of different type separately, given that models showed a substantially different balanced accuracy across course types.
We discuss the patterns in detail in what follows. 

Concerning flipped courses, it can be observed that RF and BiLSTM models differently suffer from (un)known unknowns. 
RF models were more conservative and reported a higher percentage of known unknowns than unknown unknowns. 
It follows that this type of model will require the instructor more consciously inspect students the model was not confident on. 
On the other hand, BiLSTM models were in general more confident, but this confidence was often on the wrong prediction.
It was indeed evident that BiLSTM models led to more unknown unknowns than known unknowns. 
In general, unknown unknowns were equally distributed between wrong failing and wrong pass predictions. 
Estimates were similar across the flipped courses, showing that the model impacted more than the course itself. 
Despite of the lower absolute values, RF and BiLSTM model patterns were confirmed on students attending MOOCs. 
It should however be noted that MOOCs included more students and, as an example, $2\%$ of unknown unknowns in $\texttt{OC-DSP}$ would involve $300+$ students. 
Analogously, $4\%$ of unknown unknowns in $\texttt{OC-HWTS}$ would involve $100+$ students. 

\hlbox{Findings RQ1}{
Unknown unknowns non-negligible prevalence was observed in both course types.
Flipped courses were more prone to unknown unknowns than MOOCs.
Given their comparable accuracy, RF models led to less unknown risks than BiLSTM models. 
Such risk was comparable between false fails and false passes. 
}

\begin{figure*}[!b]%
    \centering
    \hspace{0.8cm}
    \subfloat[R2 scores\label{fig:la-r2scores}]{
    \includegraphics[width=.75\textwidth, trim=4 4 4 4,clip]{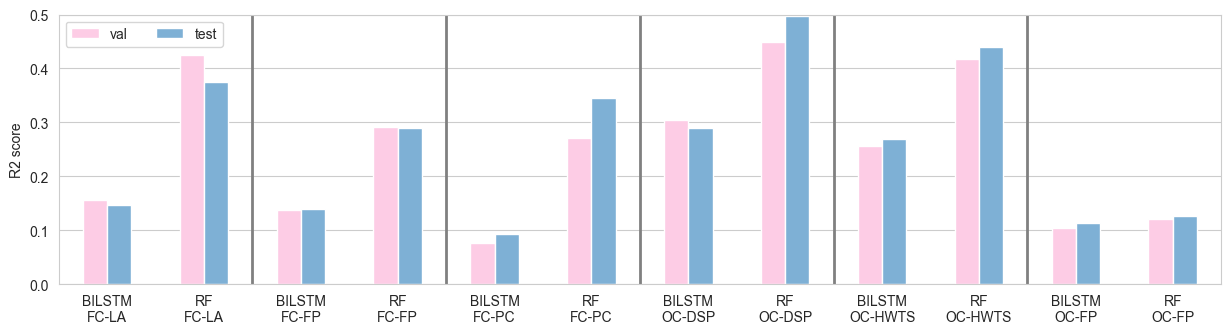}}
    \newline
    \subfloat[Highest coefficients on flipped courses\label{fig:la-coefflipped}]{\includegraphics[width=.58\textwidth, trim=4 4 4 4,clip]{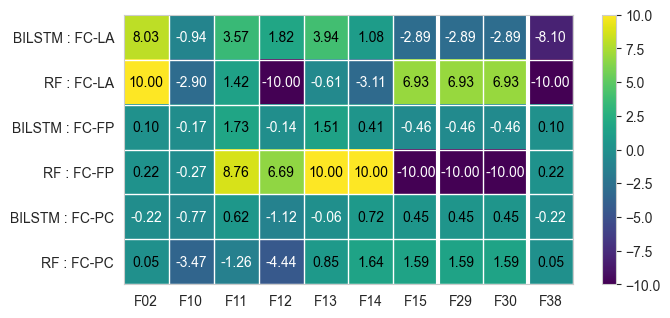}}
    \quad
    \subfloat[Highest coefficients on MOOCs\label{fig:la-coefmooc}]{
    \includegraphics[width=.32\textwidth, trim=4 4 4 4,clip]{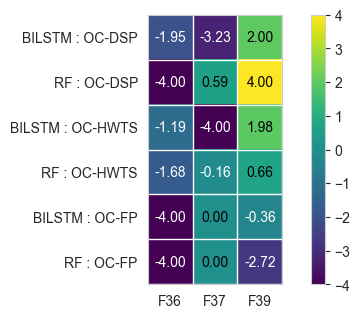}}
    \vspace{-3mm}
    \caption{\textbf{Regression Analysis}. We first collect the R2 score of the linear regression models predicting unknown unknowns (top).
    We provide the average coefficients for the indicators having an absolute value higher than $1$ on average (bottom). 
    Coefficients were clipped in the range [-10, 10] for conciseness and clarity. See Table \ref{tab:indicators} for a description of the indicators.}
    \label{fig:la}
\end{figure*}

\subsection{RQ2: Unknown Unknowns Characterization}
In a second analysis, we aimed to characterize unknown unknowns in success prediction models.
To this end, we ran the explanatory framework described in Section \ref{sec:explanatory} on top of each model and course combination. 
Figure \ref{fig:la-r2scores} collects the averaged R2 score ranging in $[0, 1]$, i.e., 
the proportion of the variance in the dependent variable predictable from the independent variable(s), under each setting.  
It can be observed that such variance was more predictable on RF models than BiLSTM models and on MOOCs then flipped courses.
Interestingly, on \texttt{OC-FP}, none of the models was able to explain reasonably well the variance in the dependent variable.
While regression coefficients might give as an initial indication of the indicators more prominent among unknown unknowns, 
the (reasonably low) R2 score confirmed our hypothesis that a supplementary model for unknown unknowns detection would also not be the most viable solution.   

Figure \ref{fig:la-coefflipped} and \ref{fig:la-coefmooc} summarize the indicators having the highest coefficient for predicting unknown unknown's membership. 
For conciseness, we limited our discussion in the paper to the indicators whose coefficient had on average a value higher than $1$ across courses of the same type. 
Concerning flipped courses, ten behavioral indicators were found important to explain unknown unknowns.
Indicators pertaining to the control dimension were more present than those related to engagement and proactivity. 
For instance, the frequency of play (F10), stop (F11), and speed events (F15), but also the total number of video (F29) and problem clicks (F30), showed a high positive weight for being unknown unknown. 
In our data set, we observed that the majority of students who passed indeed tended to perform these actions many times. 
We conjecture that the model was challenged to understand whether those actions were frequently performed 
because students were struggling or they were reflecting more on the concepts and fixing them in their mind (this is something the model was not aware of). 
Therefore, the model tended to classify as passing with a high confidence also those students who were struggling with the content. 
Other observations can be made for the frequency of stop events (F12) and the alignment with the schedule (F38), which corresponded to a very low negative weight.  
Being not aligned with the schedule would not always mean the student will likely fail the course. 
For instance, a student could have learnt consistently offline and model unawareness about this could lead to unknown unknowns. 

In MOOCs, three indicators, all pertaining to proactivity, were found relevant through our explanatory analysis. 
No overlap in the selected regression coefficients was observed between flipped courses and MOOCs. 
Content anticipation show often associated to a high regression coefficient. 
Indeed, for students eager to learn, anticipating some course content can be a positive attitude for passing. 
For other students, who were maybe just interested in seeing what would have come next, interacting with the content in advance with respect to the schedule might not be a signal for passing. Again, the model would need additional information, to be able to distinguish between cases like this one. 

\begin{table*}[!t]
\caption{\textbf{Behavioral Indicators of Unknown Unknowns}. The behavioral indicators identified as important for students being categorized as unknown unknown. For convenience, we show the IDs obtained on the full list of original features used in \cite{DBLP:conf/lats/SwamyMK22}.
}
\vspace{-3mm}
\label{tab:indicators}
\small
\centering
\resizebox{0.94\linewidth}{!}{
\begin{tabular}{@{}llll@{}}
\toprule
\textbf{Dimension} & \textbf{Indicator} & \textbf{ID} & \textbf{Description} \\ \midrule
\multirow{7}{*}{\textit{Control}}
& AvgWatchedWeeklyProp & F02 & The ratio of videos watched over the number of videos available. \\
& FrequencyEventPlay & F10 & The frequency between every Video.Play action and the following action. \\
& FrequencyEventPause & F11 & The frequency between every Video.Pause action and the following action. \\
& FrequencyEventStop & F12 & The frequency between every Video.Stop action and the following action. \\
& FrequencyEventSeekBackward  & F13 & The frequency between every Video.SeekBackward action and the following action. \\
& FrequencyEventSeekForward & F14 & The frequency between every Video.SeekForward action and the following action. \\
& FrequencyEventSpeedChange & F15 & The frequency between every Video.SpeedChange action and the following action. \\
 \midrule
\multirow{2}{*}{\textit{Engagement}} 
 & TotalClicksProblem & F29 & The number of clicks that a student has made on problems this week. \\
 & TotalClicksVideo & F30 & The number of clicks that a student has made on videos this week. \\
 \midrule
\multirow{4}{*}{\textit{Proactivity}} 
& CompetencyAlignment & F36 & The number of problems this week that the student has passed. \\
& CompetencyAnticipation & F37 & The extent to which the student approaches a quiz provided in subsequent weeks. \\
 & ContentAlignment & F38 & The number of videos this week that have been watched by the student. \\ 
 & ContentAnticipation & F39 & The number of videos covered by the student from those that are in subsequent weeks. \\ 
 \bottomrule
\end{tabular}}
    \vspace{3mm}
\end{table*}

By looking at the two model types on the same course, it results that different behavioral indicators confound different models on \texttt{FC-LA}. 
Indeed, the BiLSTM and RF models agreed on F02, F10, F11, and F38 but strongly disagreed with the other six.  
The two models weighted the behavioral indicators more similarly on the other two flipped courses. 
Regression coefficients went into this line on the MOOCs as well. 
Interestingly, none of the considered demographic attributes were deemed as important. 
In addition to being positive from a fairness perspective, this finding emphasizes the need of additional contextual variables, often non recorded in the log data neither in the university's database. 

\hlbox{Findings RQ2}{
Unknown unknowns membership was connected with certain behavioral indicators.
Control (more), engagement and proactivity (less) characterized them in flipped courses.
Concerning MOOCs, proactivity was shown to be the main dimension indicative of the models' unknown unknowns, regardless of the MOOC. 
} 

\subsection{RQ3: Unknown Unknowns Perception by Instructors}
\begin{figure*}[!t]
\centering
\vspace{-8mm}
\subfloat[Demographic distribution of our sample\label{fig:first_demo}]{\includegraphics[width=0.5\linewidth]{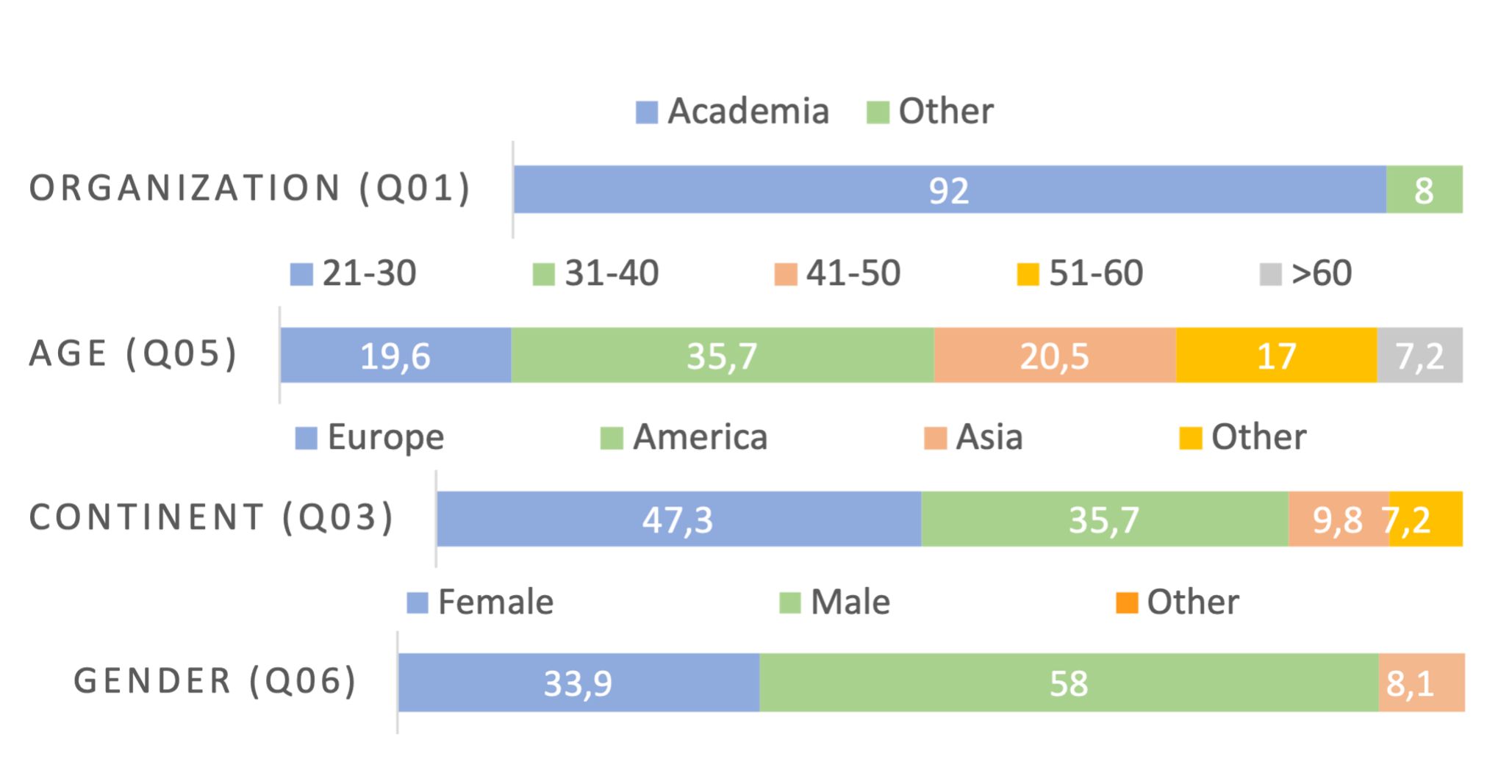}}
\subfloat[Teaching experience distribution of our sample\label{fig:second_demo}]{\includegraphics[width=0.5\linewidth]{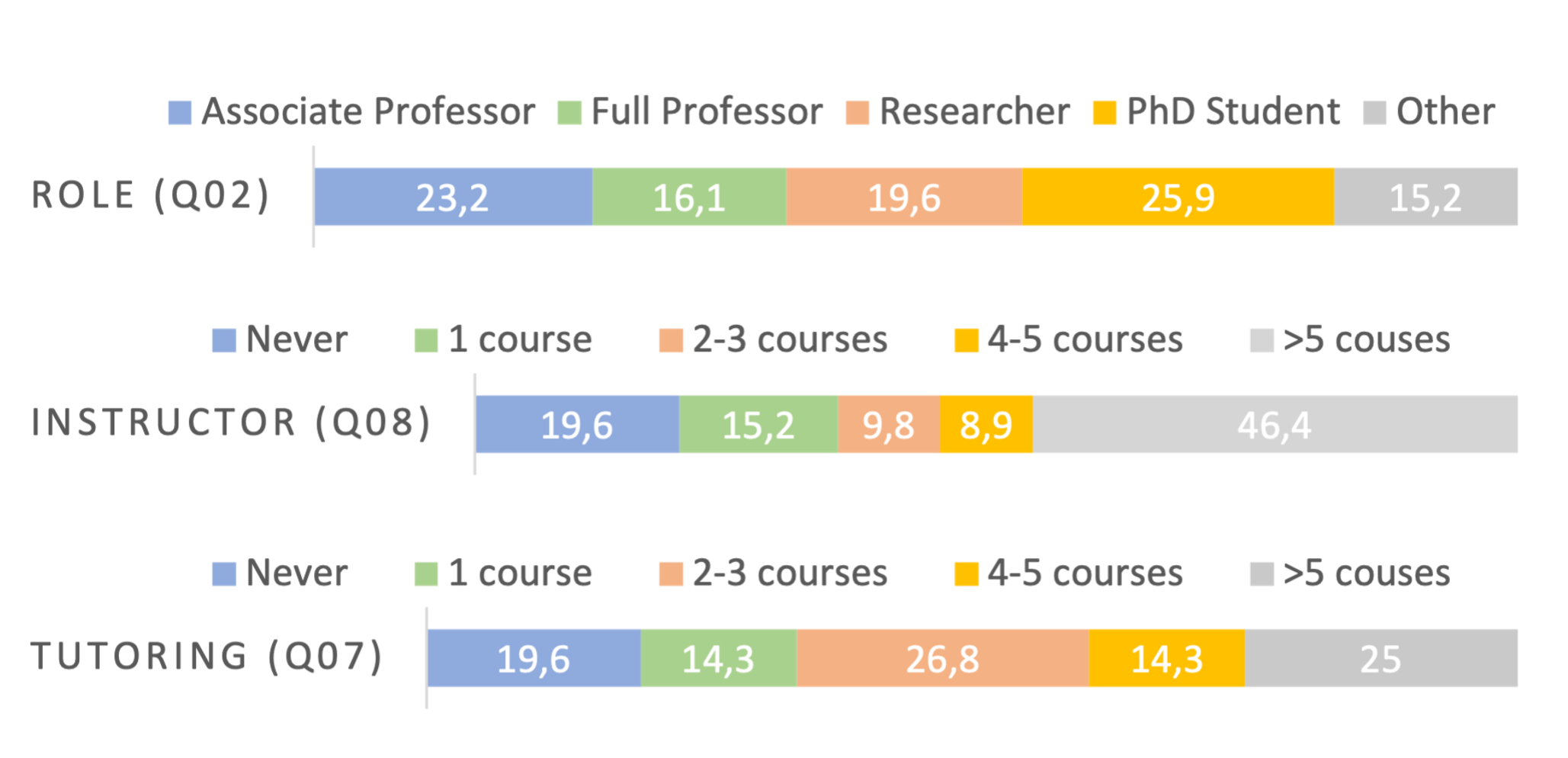}}
\newline 
\centering
\subfloat[Initial instructors' perception of the student success model\label{fig:res_1}]{\includegraphics[width=0.5\linewidth]{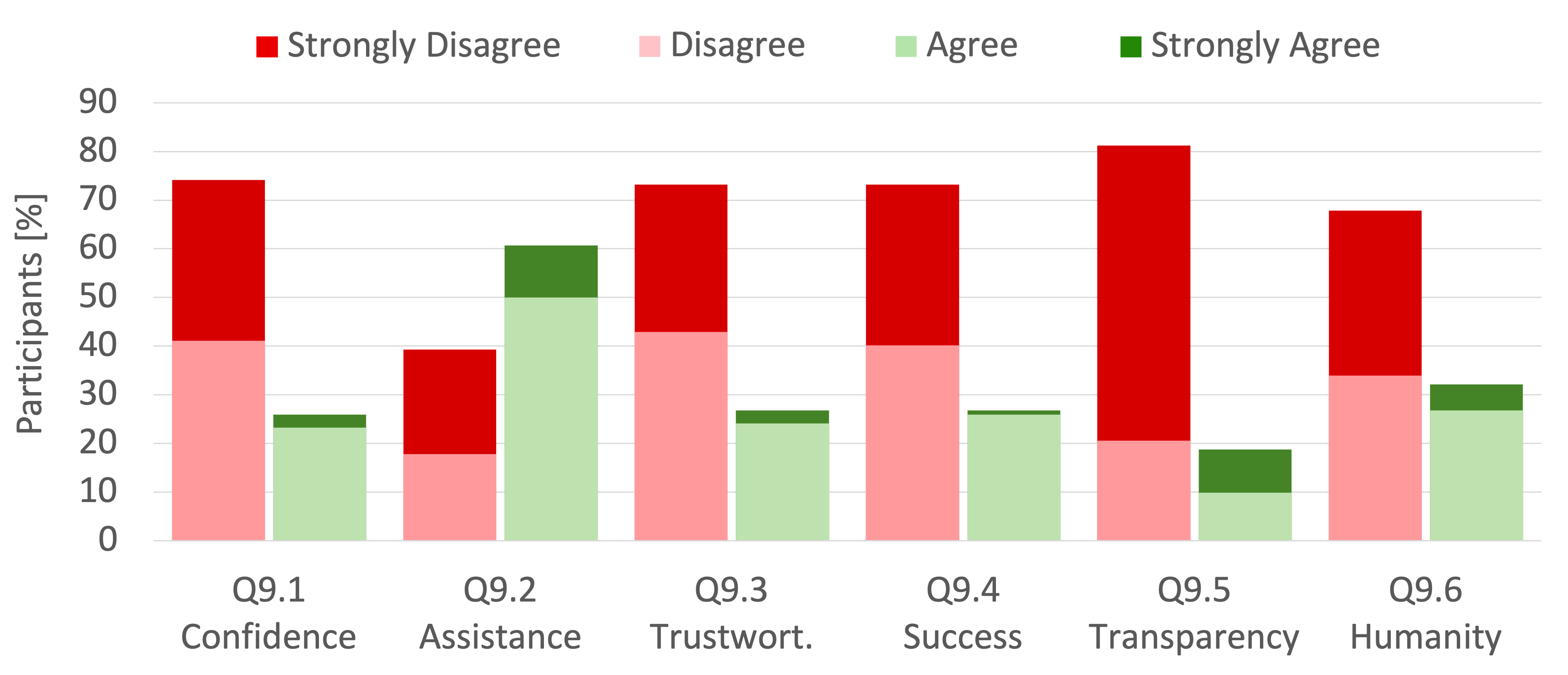}}
\subfloat[Relative change in perception w/ unknown unknown awareness\label{fig:res_2}]{\includegraphics[width=0.5\linewidth]{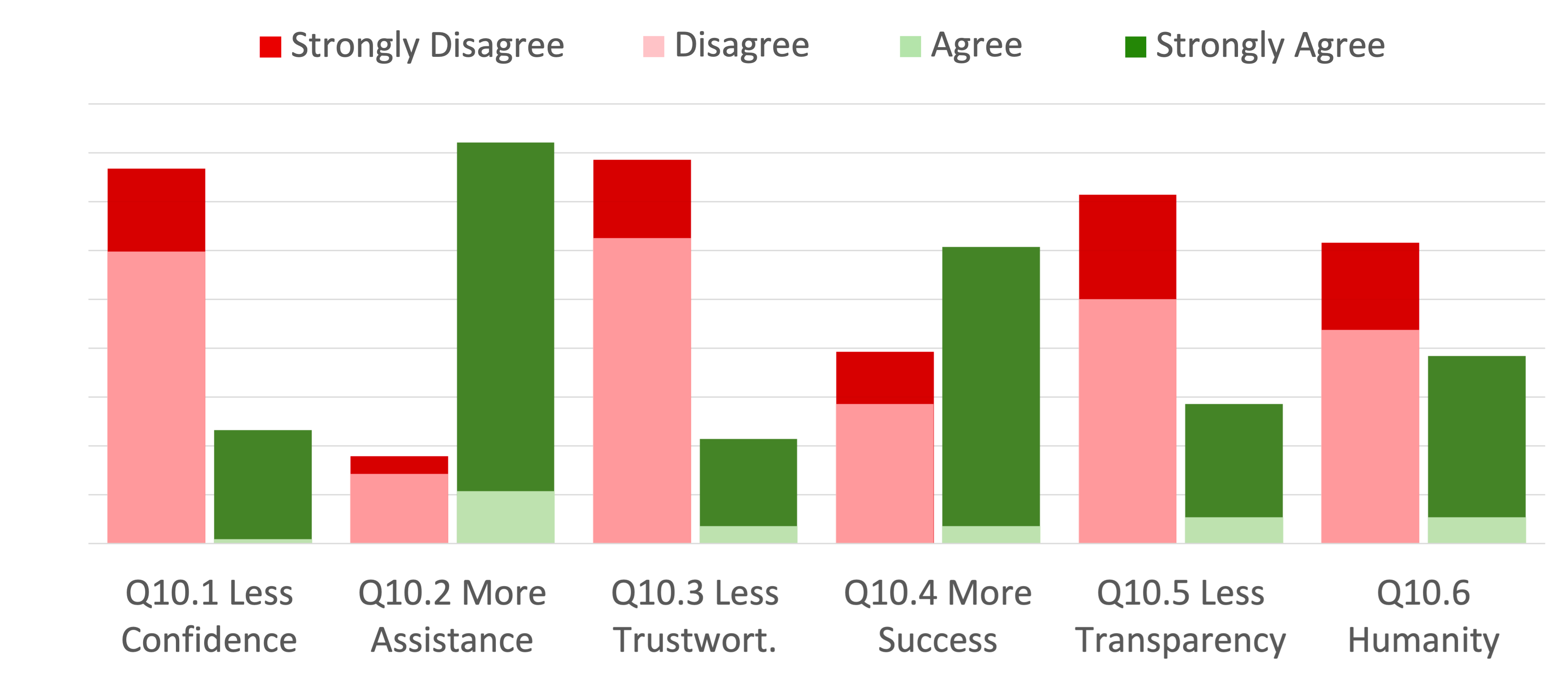}}
\newline 
\centering
\subfloat[Tutoring-wise relative change \label{fig:tutoring}]{\includegraphics[width=0.38\linewidth]{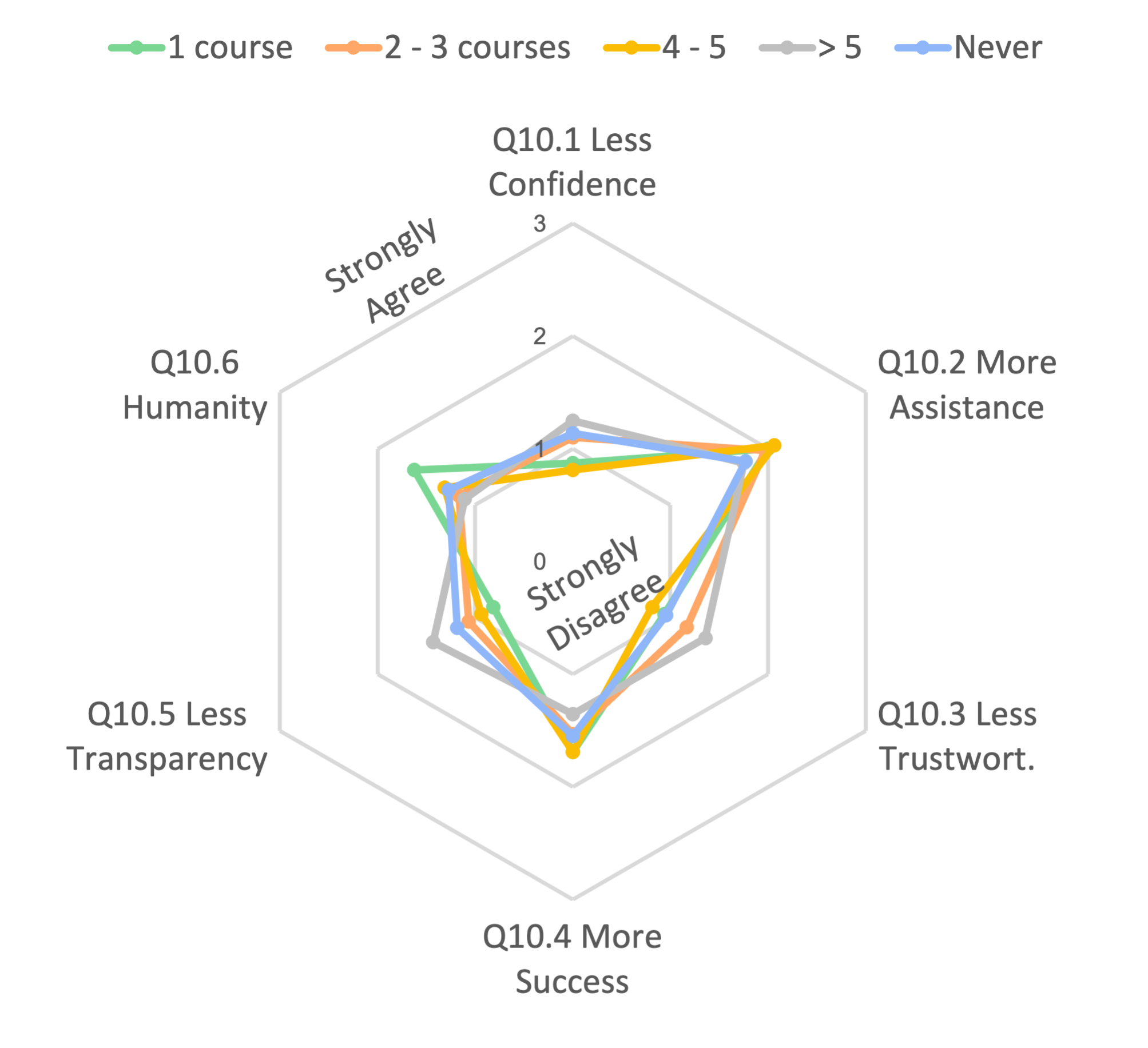}}
 \centering
\vspace{-2mm}
\subfloat[Instructor-wise relative change \label{fig:instructor}]{\includegraphics[width=0.38\linewidth]{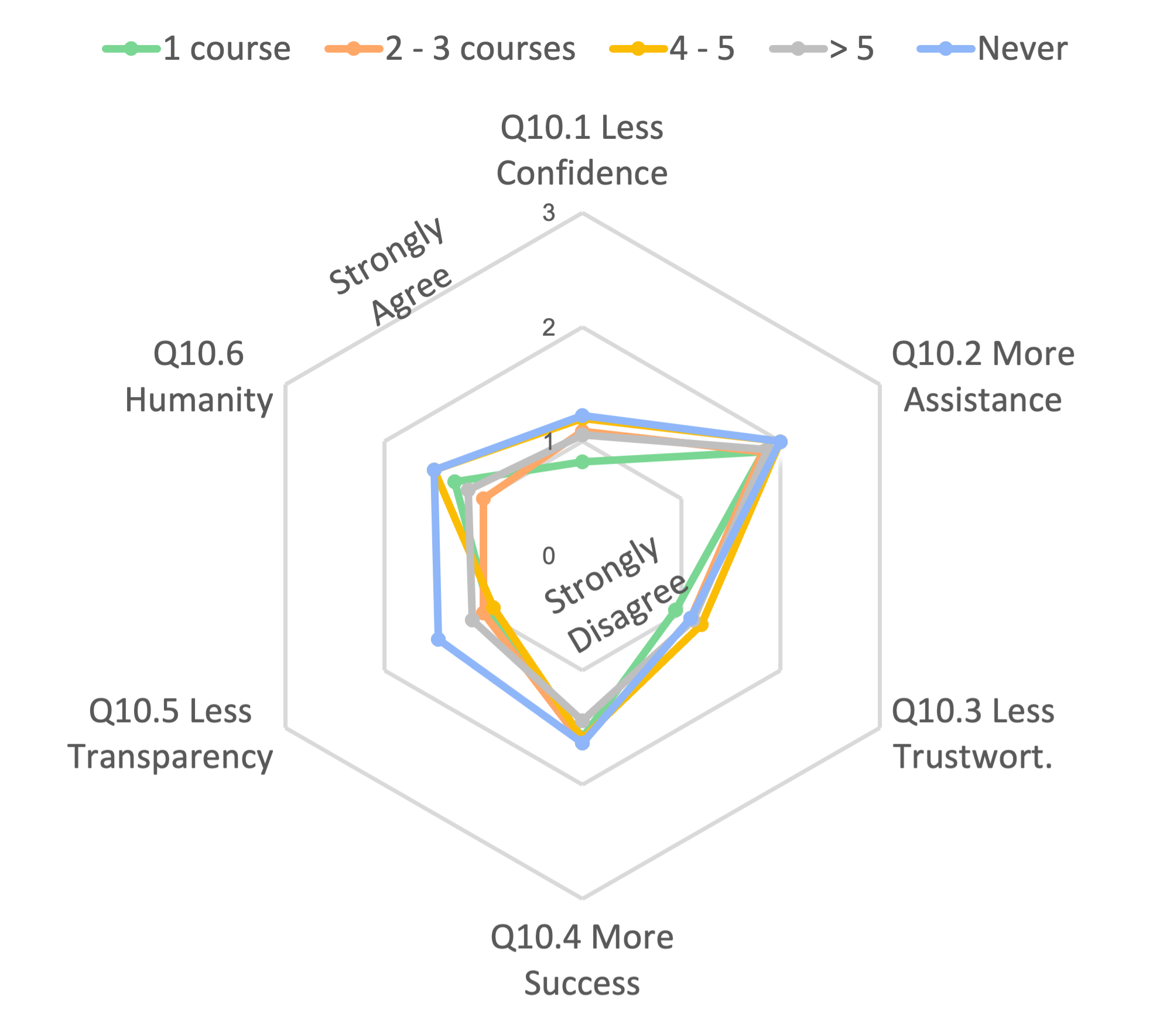}}
\caption{\textbf{Instructors' Perception}. Our sample showed a good level of diversity from both a demographic and a teaching experience perspective (a \& b).
We then summarize participants' initial perception of the exemplified student success model (c) and the relative change in their perception once they were provided with signals and explanations about unknown unknowns (d). Such relative change showed interesting patterns according to the tutoring and leading instructor experience (e \& f).}
\label{fig:teaching}
\vspace{3mm}
\end{figure*}

Finally, we were interested in understanding the instructors' perception of student success models in view of unknown unknowns awareness. 
To this end, we distributed the questionnaire and collected answers from experts worldwide as detailed in Section \ref{sec:quest}. 
Figure \ref{fig:teaching} collects a summary of the results from the questionnaire's answers.

In Figure \ref{fig:first_demo}, it can be observed that our sample mostly included participants working in an academic context (92\%)\footnote{
This aspect should be taken into account while analyzing our findings. 
This imbalance however comes from the adopted protocol, which involved a large pool of educators yet experts in the field who have recently published in the major educational conferences. 
}.
Participants were almost equally distributed across associate professors (23,2\%), PhD students (25,9\%), and researchers (19,6\%).
Our sample of participants was balanced also age-wise, despite of a slight predominance of people in the 31-40 group (35,7\%). 
With respect to gender identity, we received more responses from man (58\%)$^5$.  
Given the focus of our study, we were also interested in collecting information regarding the teaching experience (Fig. \ref{fig:second_demo}).
In our sample, we found that 19.6\% of participants never acted as tutors or instructors. 
Conversely, 25\% (46,4\%) out of the participants have served in more than five courses as a teaching assistant (leading instructor). 
Other participants reported a teaching experience between two and five courses.
Overall, our sample showed interesting patterns of diversity. 

In the second section of the questionnaire, we first investigated the instructors' perception about an example student success model (Q09).
Figure \ref{fig:res_1} highlighted that a high percentage of participants raised concerns about the example student success model. 
Participants strongly disagreed or disagreed (sum of the red-like bar values) with respect to being confident with the model predictions (74\%), trusting predictions while using them in the classroom (73,2\%), feeling of being able to use model predictions successfully (73,2\%), seeing the model predictions as transparent (81,25\%), and their reliability with respect to a recommendation from a human colleague (67,86\%). Overall, we observed a prevalence of strongly disagree or disagree answers, except for their feeling of assistance. It should be noted that our goal was merely to collect a baseline perception, and not to assess the perception of the instructors on student success models in general which, paired with the implications of our study, will lead to next steps in future work. 

In Figure \ref{fig:res_2}, we collected the instructors' perception after some signals and explanations about unknown unknowns were displayed into the user interface. 
To make the comparison possible, we directly asked participants to indicate the relative change in their perception across the six perspectives. 
To this end, we alternatively added the terms "\emph{less}" and "\emph{more}" to the same questions originally reported in Q09. 
We decided to alternate positive (more) and negative (less) statements to reduce potential noise and biases in the answers. 
Our results showed that a high percentage of participants felt more confident (76,79\% of the participants strongly disagreed or disagreed with the negative statement).
A participant for instance said that "\textit{It helps teachers understand the reasoning behind the prediction. It helps on increasing my confidence on the model.}". 
Instructors felt more assisted (82,14\% of the participants strongly agreed or agreed with the positive statement). 
For instance, a participant found that "\textit{it was possible to get more insight}". 
Trustworthiness was positively impacted as well, with 78,57\% of the participants (strongly) disagreeing with the negative statement. 
Instructors felt also more successful in using model predictions (60,71\% of the participants strongly agreed or agreed with the positive statement). 
They also thought that model predictions were not less transparent, so this perspective was positively impacted (71,43\% of the participants strongly disagreed or disagreed with the negative statement). 
However, instructors still did not truly rely on model predictions (61,61\%) as much as human recommendation.
A participant emphasized that "\textit{I would not fully trust the predictions but they could help me to identify which students that I should I pay attention to}".
To have a detailed picture, we then cross-referenced these values with the different demographic elements in Figure \ref{fig:tutoring} and \ref{fig:instructor}. 
Interestingly, a participant reported that "\textit{The major advantage with additional alerts/explanation is the added context. In absence of explanations, the model is a strict "follow or not follow" type, but with these additional alerts its helps instructors to assess/study the students learning behavior and will help drive usability of the model.}"

\hlbox{Findings RQ3}{
Being made aware of unknown unknowns information, instructors feel a higher confidence, assistance, trust, success, transparency.  
Though their perception slightly improved, instructors still did not truly rely on model predictions as much as human recommendation.
} 

\section{Conclusions and Implications}
With our experiments, 
we showed that unknown unknowns exist and vary in number and type across courses (\textbf{RQ1}).  
We then characterized unknown unknowns under the specific use cases (\textbf{RQ2}).  
Finally, we found that making instructors aware of unknown unknowns had a positive impact (\textbf{RQ3}).   
Our findings led to multiple implications. 

\vspace{1mm} \noindent \textbf{Scientific Implications}.
The collected data is usually partial and does not provide the global picture of students' behavior, skills, and needs. 
Many other variables can be hardly collected for being included in the model reasoning. 
For instance, a student might work a lot offline and still get very good exam grades.
Being the predictions dependent on the data, there is the high risk that at the moment no enough data for the model is collected.
Our study therefore calls for a more extensive data collection aimed to bridge the gap between what the models knows and what should know. 

Our study has shown that, BiLSTM models were often very confident about their predictions, but completely wrong in several cases. 
When possible, model uncertainty would be preferred to avoid misleading instructors, as in RF models.  
Understanding what is the source of such model confusion and how this knowledge can be induced into the model is urging.
Another implication is therefore the need of student success models that reduce unknown unknown cases. 

Notably, we proved that assessing model performance solely based on accuracy may introduce unknown risks. 
When a student success model is delivered, evidence on its unknown unknowns should be provided. 
This evidence could be both quantitative, by reporting for instance the percentage of unseen students resulting as (un)known unknowns,
but also qualitative, by characterizing the cases where the model is less confident but incorrect. 

Instructors will be likely to use success models as a complementary support to their personal perception. 
Human-in-the-loop approaches can be used to let the instructor and the model help each other while identifying students that require assistance. 
Unknown unknowns represent examples the instructor should reflect on while using predictions. 

\vspace{1mm} \noindent \textbf{Technological Implications}.
Our study has proven that confidence levels are not enough to prevent undesired behavior like unknown unknowns. 
Signals of their presence, though helpful, would just be triggers for further analysis of certain students. 
Indeed, pass/fail predictions alone would not be enough and more insights about learner behavior will be needed.
Behavioral patterns (e.g., late submissions) can then indicate how to counsel students.

In our work, we adopted BiLSTM models, with sigmoid and cross entropy. that could tend to push predictions towards the two sides, 0 and 1. 
Our findings show that this practice creates more unknown unknowns (more risks). 
Future work should carefully consider this aspect while selecting model parameters to reduce unknown risks.

Once unknown unknowns patterns are identified, it will be important to understand how information about them should be presented. 
Having signals and explanations has led in our study to an increase of trust in the model prediction.
Rather than a signal, showing a null state when a prediction is potentially risky would be another solution. 
Instructors might feel that the model did not have enough data to learn the likely outcome for some of the students yet. 

Besides being used for learning understanding, student success models can fuel tools to recommend instructors student requiring assistance (especially in large classes).
It could be also important to have grouped predictions, along with individual predictions, based on similarities. 
Reducing unknown risks will be even more important in these settings. 

\vspace{1mm} \noindent \textbf{Social Implications}.
In our results, we showed that raising awareness of unknown unknowns led to higher confidence, feeling of assistance, trust, and transparency.
Being trustworthiness a complex challenge in this field, our study introduces another source for improvement, complementary for instance to explanations. 
Furthermore, such models could be harmful by creating false expectations.
Being aware of unknown unknowns can help to prevent such situations. 

Finally, predicting whether a student is going to pass or fail a course is of practical utility if it is done early. 
Once all of the data is collected over the course end, the models' purpose might be just to detect at-risk students who might benefit from remedial sessions between the course end and the final exam. 
Nevertheless, we believe that the unknown unknowns might be even more evident in very early predictions, which we plan to investigate in future work. 

\vspace{1mm} Scientific, technological, and social shifts in education often go hand-in-hand.
Countering unknown unknown issues will be essential to further strengthen the reliability of emerging student success models in real-world education.

\begin{acks}
This project was substantially co-financed by the Swiss State Secretariat for Education, Research and Innovation SERI. 
Roberta Galici gratefully acknowledges the University of Cagliari for the financial support of her PhD scholarship.
\end{acks}

\bibliographystyle{ACM-Reference-Format}
\bibliography{sample-base}

\end{document}